# 3D Contouring for Breast Tumor in Sonography

*Dar-Ren Chen, MD[†], Yu-Chih Lin[*] Yu-Len and Huang, PhD[*]*

[*] Department of Computer Science, Tunghai University, Taichung, Taiwan

[†] Comprehensive Breast Cancer Center, Laboratory of Cancer Research, Changhua Christian Hospital, Changhua, Taiwan

**Address correspondence and reprint requests to**:

Professor, Yu-Len Huang
Department of Computer Science
Tunghai University
No.1727, Sec.4, Taiwan Boulevard, Xitun District, Taichung 407, Taiwan
Tel: (886)-4-23590121 ext. 33800
Fax: (886)-4-23591567
E-mail: ylhuang@thu.edu.tw

# Abstract

Malignant and benign breast tumors present differently in their shape and size on sonography. Morphological information provided by tumor contours are important in clinical diagnosis. However, ultrasound images contain noises and tissue texture; clinical diagnosis thus highly depends on the experience of physicians. The manual way to sketch three-dimensional (3D) contours of breast tumor is a time-consuming and complicate task. If automatic contouring could provide a precise breast tumor contour that might assist physicians in making an accurate diagnosis. This study presents an efficient method for automatically contouring breast tumors in 3D sonography. The proposed method utilizes an efficient segmentation procedure, i.e. level-set method (LSM), to automatic detect contours of breast tumors. This study evaluates 20 cases comprising ten benign and ten malignant tumors. The results of computer simulation reveal that the proposed 3D segmentation method provides robust contouring for breast tumor on ultrasound images. This approach consistently obtains contours similar to those obtained by manual contouring of the breast tumor and can save much of the time required to sketch precise contours.

# Introduction

Breast cancer is a tumor emerged from growing mammary epithelial cells or lobular, they lost control because of the growth of cancer cells and then will invade and damage nearby tissues and organs or metastasis to other organs through the blood or lymphatic system. However, breast is rich in blood vessels, lymphatic vessels and lymph nodes, breast cancer cells tend to spread to other organs easily. Breast cancer is one of the most frequently occurring cancers in women. According to the American Cancer Society (ACS) statistics, the prevalence of breast cancer rises up every year [1]. As medical technology progress continuously and personal health concerns increase constantly, making the breast cancer death rate decreased year by year. Thus early detection and early treatment for breast cancer would reduce breast cancer mortality.

Medical ultrasound is certainly convenient and safe tool for diagnosing breast tumors, particularly palpable tumors. Modern ultrasound equipment performs real-time high-resolution imaging without the use of ionizing radiation and it is relatively inexpensive and portable. Ultrasound equipment becomes the essential diagnostic tools for hospitals. Breast ultrasound is sometimes used to evaluate abnormal findings from a screening or diagnostic mammogram or physical exam. Ultrasonic images are markers for the early detection of some breast cancers. The use of ultrasound for the early detection of some breast cancers can support in differentiating between benign and malignant lesions by analyzing the homogeneity of an internal echo.

Computer-aided diagnosis (CADx) system could assist physicians by extracting useful characteristics of the tumor in ultrasound image as a basis for diagnosis [2, 3]. Three-dimensional (3D) ultrasound images of the breast cancer screening have acquired attention. 3D ultrasonography provides similar imaging quality to conventional two-dimensional (2D) ultrasound and also has the multi-sectional capability to reveal anatomical features what are not originally possible on a 2D system. The volume differences between malignant and benign breast tumors is considered a useful characteristic for identifying malignant breast tumors [4]. Thus a correct diagnosis might require accurate estimation of the total volume of the tumor and tumor contour. In conventional 2D ultrasound, the expert must mentally construct a 3D impression of tumor based on many 2D ultrasound images.

The accuracy of CADx is related to the location and extent of tumor, the segmentation of the tumor has a direct impact on the accuracy of diagnosis. Contour of breast tumor can be sketched in manual, semi-automatic, or fully automatic manner [5-8]. The manual segmentation method is not suitable for 3D ultrasound images. A 3D sonogram in one scan can take over hundreds of 2D image slices, which required plenty of time to sketch tumor contour manually. Thus automatic contouring which provided the similar contour with manual sketch of the breast tumor might assist physicians in making correct diagnoses [9]. However, ultrasound images comprise a lot of spots, noise, shadow and tissue texture; in addition, compared with the benign tumors, the edge of malignant tumors seemed more irregular, crushing and splitting, leading to the automatic segmentation method becomes a complicated work [10]. Most of segmentation methods on 3D ultrasound images utilized semi-automatic way to make up for technical shortcomings. The operations of complexity and efficiency, the stable results of the execution are also worth consideration factors [11].

In computer vision, image segmentation is the process of partitioning a digital image into multiple sections. The purpose of image segmentation is to simplify or change the representation of the image, making the image easier to understand and analyze. There were many notable approaches for ultrasound image segmentation, such as clustering-based image segmentation and region-based segmentation methods.

Clustering-based image segmentation methods can be viewed as methods of minimizing the intensity distance between each pixel and cluster center. K-means clustering [12, 13], which is a clustering-based segmentation method, segmented image into K clusters through repeated iteration until no changes and then the method can create the effect of clustering segmentation. However, the K-means clustering data is also sensitive to noise and isolated points. This method must set the number of clusters, and excessive dependence on the initial value.

Region-based segmentation according to the similarity of pixels, gathered the pixels into one area by exploiting the grayscale brightness, color and material. Equivalent to the whole image is divided into a block. Watershed transformations [14, 15], which is a popular region-based segmentation method, exercises the concept of topography and treats digital image as a topographic plane. The pixel with highest value is the highest point on the terrain, the larger the change of the values indicates a steeper terrain. However, watershed transformation is very sensitive to noise, which would cause over-segmentation to obtain inaccurate contours [16]. Active contour model (ACM) algorithm [17], also known as snake algorithm, mainly used in two-dimensional image analysis and machine vision image boundary detection, the purpose is to find the object boundary. Although active contour can be more accurately search the boundary of the object, the drawback is the need to provide the initial boundary and works on single target only. Moreover, the ACM-deformation procedure is very time-consuming.

The aim of this study was to develop an efficient segmentation scheme for 3D breast ultrasound imaging. This study describes a contour initializing procedure by using 3D region growing techniques, as part of the proposed contouring method, to generate formal initial contours for breast tumors. Level set method (LSM), an efficiently developed deformable model that surpassed ACM in segmentation performance, is utilized to conserve the time required to sketch a precise contour. The proposed method performed 3D version of the LSM to extract contours of a breast tumor from 3D ultrasound imaging with very high stability. This study then compared the automatic breast contouring results with those sketched by an expert radiologist.

## Materials and Methods

**Data Acquisition**

This study collected ten benign and ten malignant breast tumor patients aged from 20 to 78 years old (mean age 42 years). The entire database was approved by the institutional review board and ethics committee at Changhua Christian Hospital, Taiwan. Informed consent was obtained from all patients. If the patient had multiple breast lesions, the biggest lesion was included in the study. Approximately 120 to 280 ultrasound images of each case and the average image size is 366×216 pixels. Sonographic examinations were done by using 3D power Doppler ultrasound with the high definition flow (HDF) function (Voluson 730, GE Medical Systems, Zipf, Austria). A linear-array broadband probe with a frequency of 6–12 MHz, a scan width of 37.5 mm, and a sweep angle of 5° to 29° to obtain 3D volume scanning was used. Physician kept a fixed sweep angle of 20° and power Doppler settings of mid frequency, 0.9 kHz pulse repetition frequency, -0.6 gain, and 'low 1' wall motion filter in all cases. All obtained images were stored on the hard disk and transferred to a personal computer using a DICOM (Digital Imaging and Communications in Medicine) connection for image analysis.

**The Proposed Segmentation Method**

The quality of initial contour not only influences to the final segmentation result, but also relates to computation time of segmentation. The proposed method utilized the 6-neighbors 3D region growing method to generate initial contour. Morphological closing operator [18] was used to refine the extracted contour which generated by 3D region growing procedure. The closing operator could eliminate small holes, cracks, and elongated pit; and then filled the gap on the contour to smooth the boundary. Finally, this study performed LSM algorithm by using the extracted initial contour to obtain the desired segmentation results. Figure 1 shows the flowchart of the proposed method.

*Initial Contouring*

Region growing technique [19, 20] is a simple and effective method for extracting initial contours. The procedure starts with a set of seed points and then grouped neighboring voxels or sub-regions which provided similar properties. To determine that the seed surrounding pixels whether to have similar characteristics with the seed, like gray-scale values, joints and colors. If the surrounding voxels had similar characteristics with the seed, the surrounding voxel was accepted into the same region. 3D region growing method [21] was used to separate the breast tumor region from background.

To extract initial contour of the tumor designate in 3D ultrasound imaging, a physician experienced in breast ultrasound examination defined and manually selected seed points which approached tumor center. The 3D regional growing procedure in this study performed six 6-neighboring model (6-connectivity) [22] along the selected seed point, i.e. front, rear, left, right, upper and lower directions, to find a similar part with similar characteristics (see Fig. 2). The intensity property was used to include a voxel in either region if the absolute difference in intensity between a voxel and the seed point is less than a predefined property threshold *T*. Adjacent voxels that satisfied the intensity property for the sub-region was assigned into tumor region A. If difference between any of the adjacent voxels and the seed point was less than the threshold *T*, the proposed method put the adjacent voxel incorporated into the same region with the seed point and re-calculated mean intensity of A. Then the new incorporated voxel was used as a new starting point to look for its six adjacent voxels. The entire procedure was repeated until no further changes occurred. Due to 6-neighboring model is faster than the traditional 26-neighboring model, the proposed procedure required a small amount of computation time to generate a reasonable initial contour of breast tumor in 3D ultrasound imaging.

*3D Refining Segmentation*

The proposed method performed 3D LSM algorithm to obtain refined contour of breast tumor. The LSM is a numerical technique for computing and analyzing the curve propagation [23]. The basic idea of LSM is that the plane closed curve expressed as a 3D continuous function surfaces which having the same function value curve $\varphi(x, y)$ is usually $\varphi = 0$, known as the zero level-set, and the $\varphi(x, y)$ is also called level-set function. The LSM is a way to denote active contours [24]. The zero level-set contour of the function is defined by

$$C \{(x, y) | \varphi(x, y) = 0\}. \qquad (3)$$

Moreover, inside region and outside region of the curve is defined as

$$\begin{cases} \varphi(x, y) > 0 & \text{outside region} \\ \varphi(x, y) = 0 & \text{contour} \\ \varphi(x, y) < 0 & \text{inside region} \end{cases}. \qquad (4)$$

In the LSM, $\varphi = 0$ represents the desired contour, $\varphi < 0$ is the internal region of the contour, and $\varphi > 0$ is the external region of the contour. Figure 3 shows that the value of level-set function was used to outline the desired contour.

Take a closed curve as a boundary, the entire plane is divided into two areas: the external and internal regions of the curve. In the plane defined distance function $\varphi(x, y, t) = \pm d$, where $d$ is the point $(x, y)$ to the shortest distance of the curve, function symbols depending on the point in the curve of the internal or external, generally defined curve interior points the distance is negative, $t$ represents time. At any moment, the point on the curve is the zero point of distance function value, i.e. a function of distance to the zero level set. Level-set method not only can handle a sharp endpoint and concave angle, but also can be automated to change the topology. There are two main advantages to represent the active contour by level set method. First, it can easily represent complicated contour changes, for example when the contour splits into two or develops holes inside. Second, we can easily know whether a point is inside or outside the contour by checking the $\varphi$ value.

This study utilized Chen-Vese method [25] which the level-set function $\varphi$ is updated as

$$\Delta \varphi = \delta(\varphi) \left\{ \mu \cdot div\left(\frac{\nabla \varphi}{|\nabla \varphi|}\right) - \lambda_1 (u_0 - c_1)^2 + \lambda_2 (u_0 - c_2)^2 - v \right\},$$

(5)

where $u_0$ is the original image, $c1$ and $c2$ are the average gray level intensity in $\varphi>0$ region and $\varphi<0$ region, respectively. Moreover, $\mu$ and $v$ is length penalty and area penalty, respectively. The fit parameters $\lambda_1$ and $\lambda_2$ is used to adjust the weights of terms. Dirac delta function $\delta$ is performed to make the contour smooth and eliminate some small isolated regions. The second and the third terms are the driving force term which drives contour to move towards the equilibrium point. In the proposed contouring method, the original ultrasound image is first processed by the 3D region growing. After the initial contouring, the 3D level set method is performed to segment the tumor.

**Contour Evaluation**

An experienced physician who was familiar with breast ultrasound interpretations manually determined 3D contours of the tumor by using two modes. The entire manual sketching (EMS) mode denoted physician manually sketched 2D contour on each slice of a tumor. The obtained contours from the EMS mode was performed as ground truth. The partial manual sketching mode denoted the virtual organ computer-aided analysis (VOCAL) [26-28] scheme within 4D View software (GE Medical Systems, Zipf, Austria) was performed to obtain an approximated 3D contour. The VOCAL scheme estimates 3D contours by a selectable degree of rotation. This study adopted a very common rotation degree 30º, the six preliminary tumor contours in 0°, 30°, 60°, 90°, 120° and 150° slice images were manually sketched. Figure 4 represents the tumor contour manually sketched with 30° rotation. The VOCAL mode utilized the six extracted tumor regions to build a 3D interested region volume.

The obtained contours from two modes were saved in files for comparison with the automatically generated contours. Four practical similarity measures [29], the similarity index (SI), overlap value (OV), overlap fraction (OF) and extra fraction (EF) between the manually determined contours and the automatically detected contours were calculated for quantitative analysis of the contouring results. REF represents the results depicted by the EMS mode, and SEG indicates the results from the VOCAL mode or the proposed segmentation method. Overlap area denotes the area covered by SEG and REF, extra area denotes the false positive area and missing area denotes false negatives area.

Figure 5 illustrates the relationship between the SEG and REF. When SI, OV and

OF are close to 1, and EF computation is close to 0, it means that the contours generated by automatic segmentation is similar to the manual contours by physician. The overlap denotes the area of the intersection of the reference and the automated segmentation. The SI expresses the similar degree between SEG and REF areas. The OV is identical to the Jaccard index, also known as Intersection over Union (IoU). The SI, OF, OV and EF are defined as

$$SI = \frac{2\times(REF\cap SEG)}{REF+SEG} \times 100\%, \tag{5}$$

$$OF = \frac{REF\cap SEG}{REF} \times 100\%, \tag{6}$$

$$OV = \frac{REF\cap SEG}{REF\cup SEG} \times 100\% \text{ and} \tag{7}$$

$$EF = \frac{\overline{REF\cap SEG}}{REF} \times 100\%. \tag{8}$$

## Results

This study utilized the similarity measures to evaluate the tumor contours generated by the proposed segmentation method and the VOCAL method with 30º rotations. The simulations totally evaluated 20 cases with manual sketched contours form EMS mode (including 10 benign breast tumors and 10 malignant ones) to test the accuracy of the proposed contouring method.

In this study, the region growing threshold $T$ was experimentally set to 5.0, and the size of structure element for closing operator was $20 \times 20$. 3D Gaussian blur filtering with sigma value 1.5 was performed to diminish noises in imaging before level-set segmentation. And then we employed the level set method to achieve a precise segmentation for the tumor. With the curvature parameters length penalty $\mu = 0.2$, area penalty $\nu = 0$, fit weights $\lambda_1 = \lambda_2 = 1$, the proposed level scheme obtained a stable and the highest accuracy.

Table 1 shows contouring performance of the proposed method and VOCAL method. The average of the measures (SI, OV, OF, EF) that determined by the proposed method and the VOCAL method were (0.85, 0.92, 0.75, 0.13) and (0.81, 0.84, 0.69, 0.34), respectively. The simulations were made on a single CPU Intel® Xeon® Processor E3-1225 v5, 3.30 GHz personal computer with Microsoft Windows 10 professional operating system and the program development environment was MATLAB (R2016.a) software (The MathWorks, Inc., Natick, MA). Average execution time for each case was 25.3 seconds. As this scheme is an offline diagnostic application, these segmentation times are clinically acceptable.

Figure 6(a) is the contour of a benign tumor which drawn manually by a doctor. Figure 6(b) is the result obtained by the proposed method. Figure 6(c) is contour result obtained by using VOCAL method. Besides, Fig. 7(a) is the contour of a malignant tumor which drawn manually by a doctor. Figures 7(b) and 7(c) are the obtained result by the proposed method and VOCAL method, respectively. Figure 8 shows the detail results of the assessment of the proposed method and the VOCAL method. The proposed method clearly yielded contours that are more similar to those manually sketched than that of obtained by VOCAL method. From the segmentation results, only a small number of cases might generate an undesired segmentation.

## Conclusion

Today, in many diagnostic modalities of breast cancer, advantages of ultrasound are images-cost, easy to operate, non-radioactive, immediate angiography and non-invasive. Physicians generally utilized ultrasound images in tumor diagnosis and testing, moreover clinicians take it to test tumor biopsies to determine tumor benign and malignant. If a computer-aided system was useful to correct depiction of tumor contour, it would be help on improving physician diagnosis of benign and malignant tumors of the correct rate.

This study proposed a fast and high accuracy automatic 3D breast tumor image segmentation method. The proposed method performed the region growing produce to obtain the initial outline and get the final contouring results by using level-set method. This work kept away from the problem of level-set algorithm which needs a large amount of computation in 3D space, as well as resolved its problem of excessive dependence of initial contour in image segmentation.

According to the experimental 10 benign cases and 10 malignant cases showed that the methods used in this study can effectively cutting ultrasound imaging of tumor contours, its average value of SI can reach more than 85%; compared to the VOCAL method of the EF average 34%, the proposed method can significantly reduce the value of the EF to 12%. For tumor contour cutting error has significantly improved. In our study, each case computing time required is approximately 25 seconds. Compared to other method of cutting, usually need several minutes. Our proposed method has significant advantages, and for medical diagnostic applications have more practical assistance.

The proposed method was fully automated 3D breast tumor contour cutting. User only needed to select a region growing seed point, you will get breast cancer 3D contour cutting. Compared VOCAL method requires manual depicting six ultrasound images of the difference of tumor contours, and physicians need to manually describe each one tumor contour method of rapid and convenient, which would help doctors in the diagnosis of convenience.

Due to the reduced time required for the program operation, the proposed method does not carry out pre-processing for the image, which causes the accuracy rate resulting of breast tumors contour will be decreased. Therefore, the future will face in combination with other effective filter to reduce noise, without increasing too much computation time of the situation and effectively enhance tumor contour cutting accuracy.

# Acknowledgement

The authors would like to thank the Ministry of Science and Technology, Taiwan for financially supporting this research under Contract No. MOST 106-2221-E-029-029.

# List of Tables

# List of Figures

Table 1. The contouring evaluations of the proposed method and VOCAL method using the similarity measurements (average).

| Pathology proven result | | Benign case (10 cases) | Malignant case (10 cases) | Whole database (20 cases) |
|---|---|---|---|---|
| SI | The proposed method | **84.67%** | **85.66%** | **85.17%** |
|    | VOCAL mode | **80.81%** | **81.91%** | **81.36%** |
| OF | The proposed method | **87.66%** | **96.08%** | **91.87%** |
|    | VOCAL mode | **82.40%** | **85.79%** | **84.09%** |
| OV | The proposed method | **73.75%** | **75.34%** | **74.54%** |
|    | VOCAL mode | **68.36%** | **69.47%** | **68.91%** |
| EF | The proposed method | **11.65%** | **13.88%** | **12.76%** |
|    | VOCAL mode | **28.41%** | **38.88%** | **33.64%** |

SI: similarity index; OF: overlap fraction; OV: overlap value; EF: extra fraction

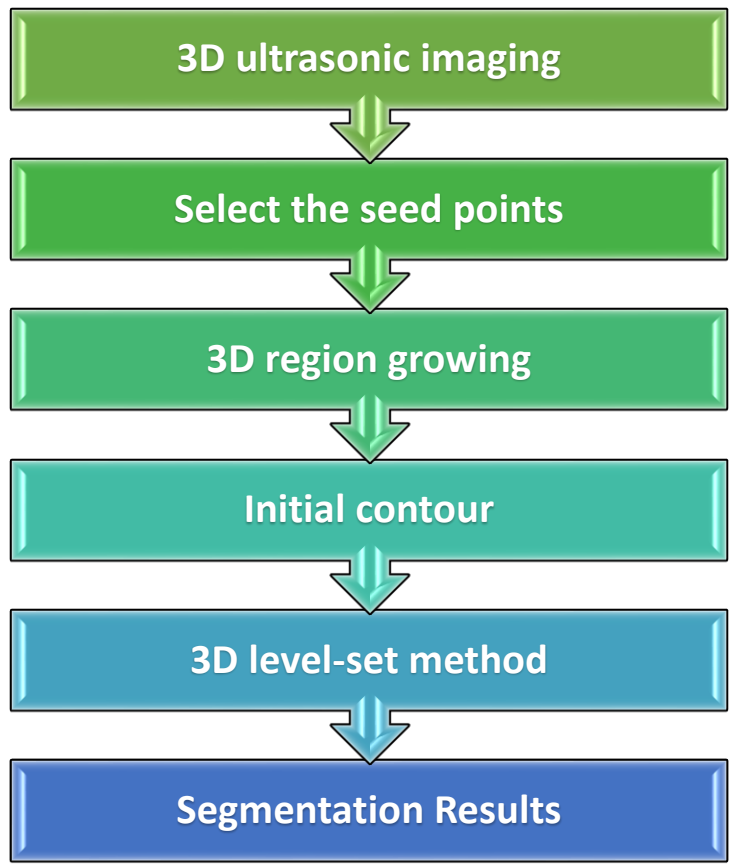

Figure 1. Flowchart of the proposed method

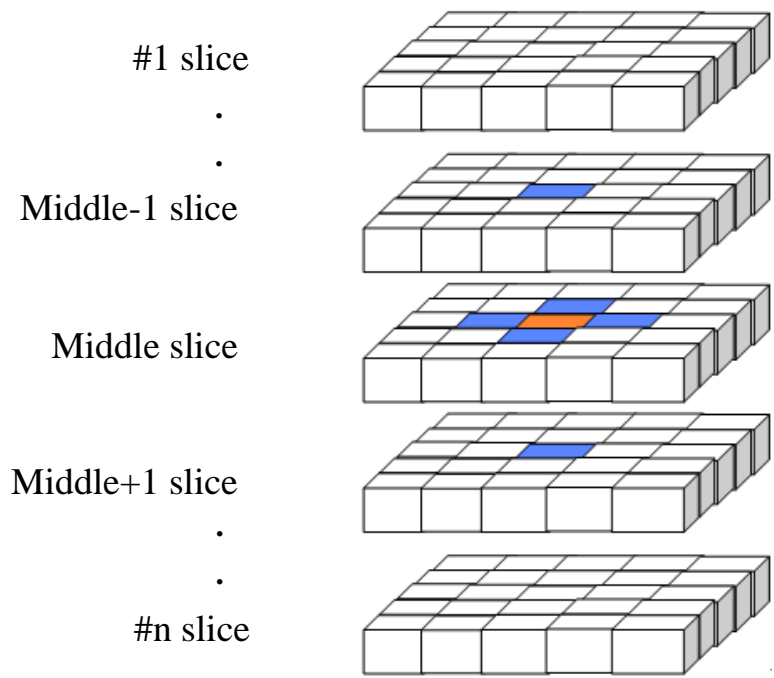

Figure 2. Neighbors (the blue voxels) within the 6-connectivity

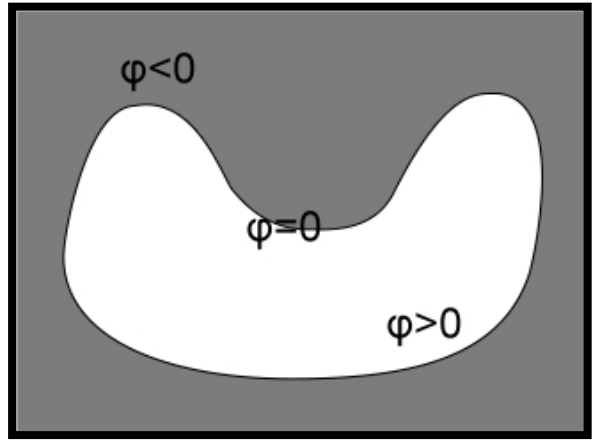

Figure 3. Illustrations of the level-set method

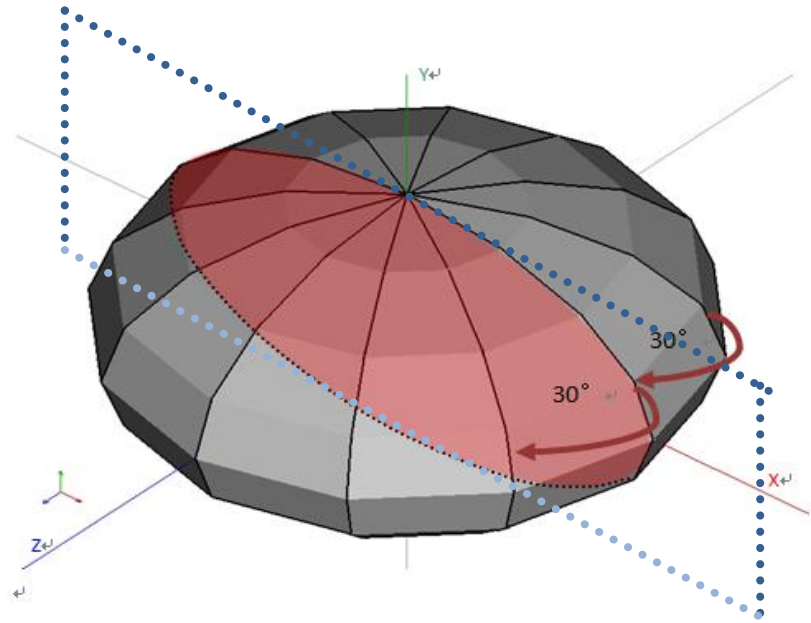

Figure 4. The tumor contour which manually sketched with 30° by export

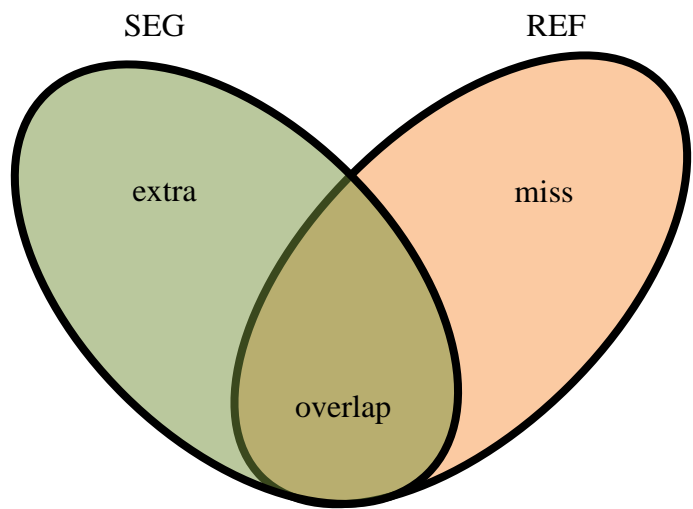

Figure 5. Schematic diagram evaluation of contour

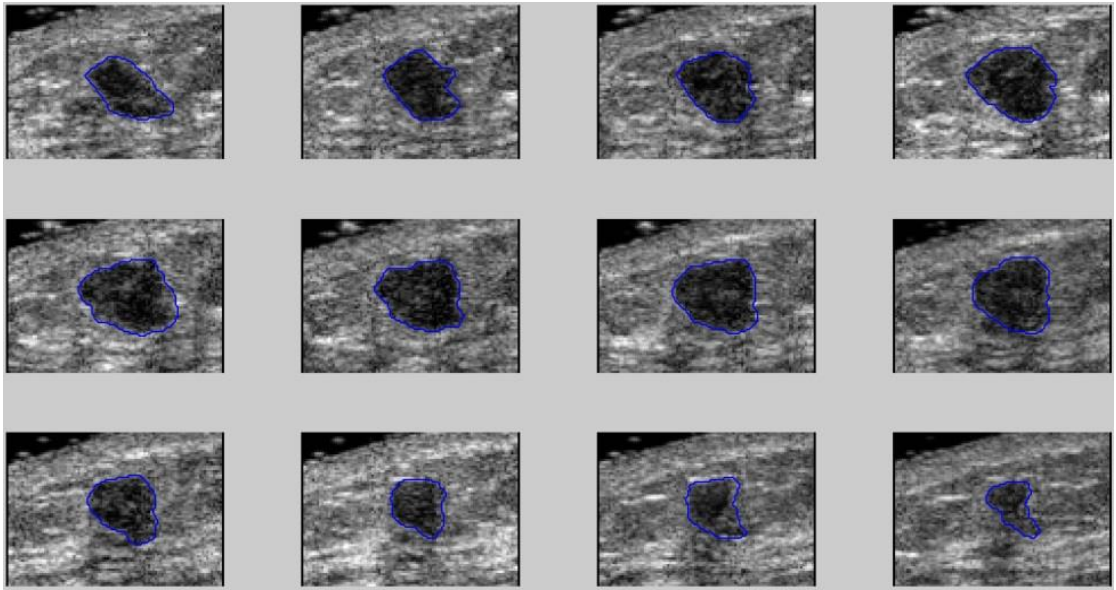

(a)

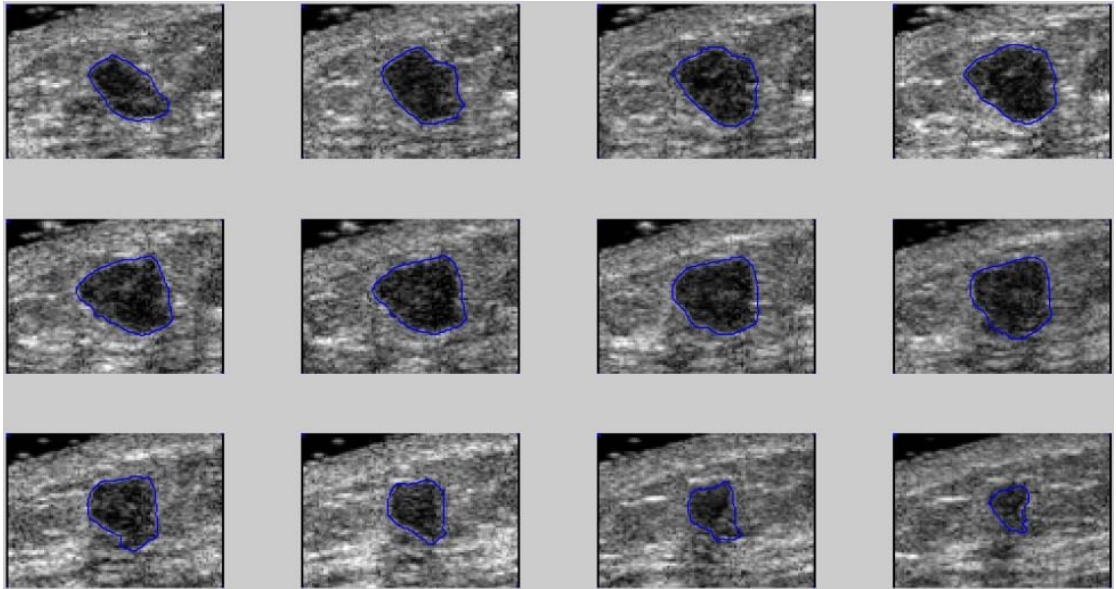

(b)

Figure 6 Results of contour segmentation with a benign case: (a) is the benign tumor contour that drawn manually by a doctor; (b) is the result of obtained benign tumor contour segmentation by our proposed LSM method; (c) is a benign tumor contour results obtained by using the VOCAL method (continued)

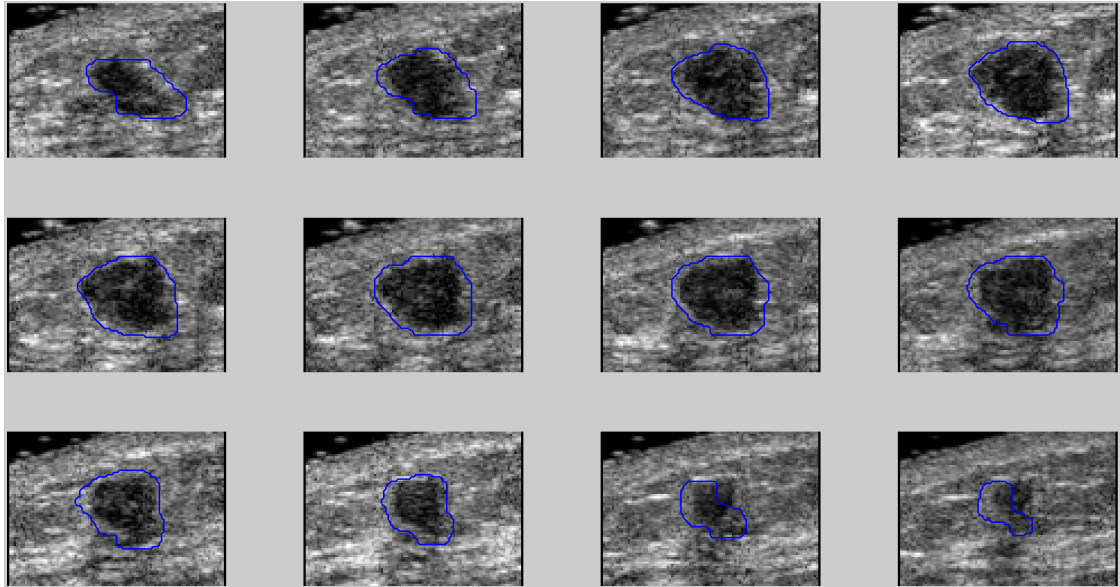

**(c)**

Figure 6. Results of contour segmentation with a benign case: (a) is the benign tumor contour that drawn manually by a doctor; (b) is the result of obtained benign tumor contour segmentation by our proposed LSM method; (c) is a benign tumor contour results obtained by using the VOCAL method

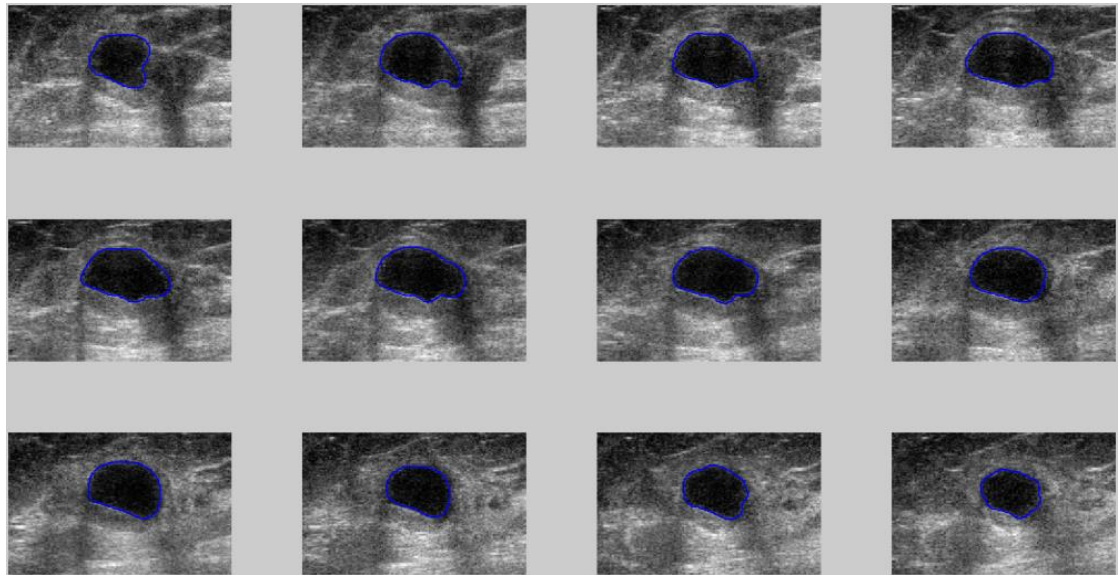

(a)

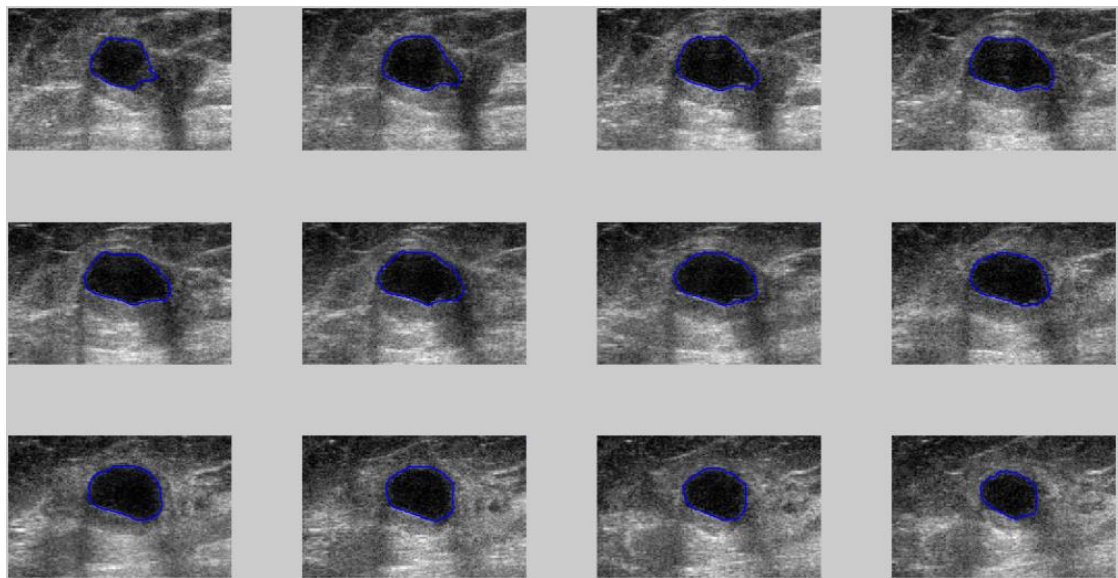

(b)

Figure 7. Results of contour segmentation with a malignant case: (a) is the malignant tumor contour that drawn manually by a doctor; (b) is the result of obtained malignant tumor contour segmentation by our proposed LSM method; (c) is a malignant tumor contour results obtained by using the VOCAL method (continued)

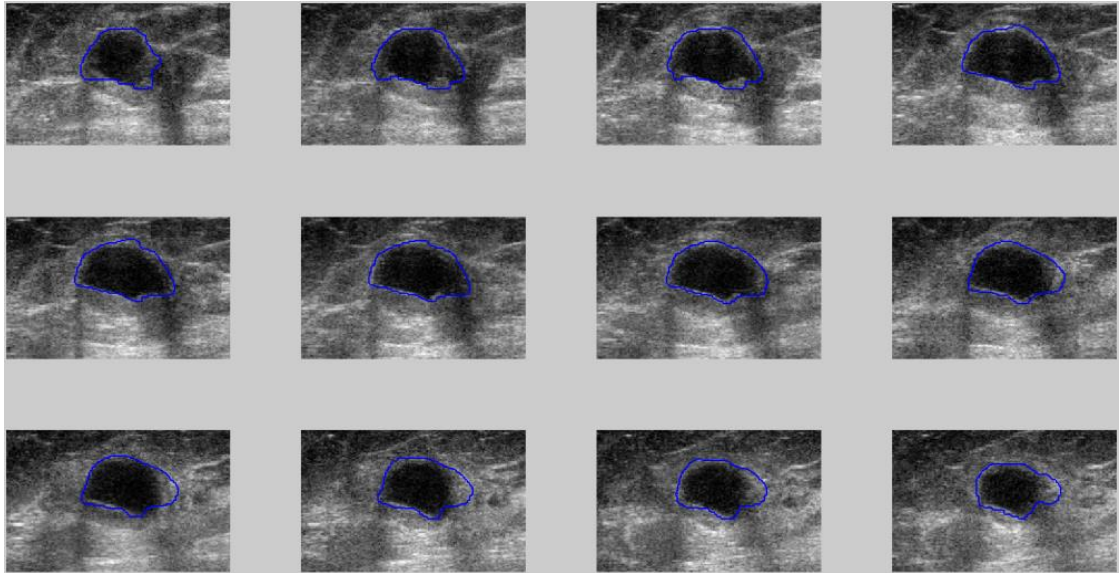
(c)

Figure 7. Results of contour segmentation with a malignant case: (a) is the malignant tumor contour that drawn manually by a doctor; (b) is the result of obtained malignant tumor contour segmentation by our proposed LSM method; (c) is a malignant tumor contour results obtained by using the VOCAL method

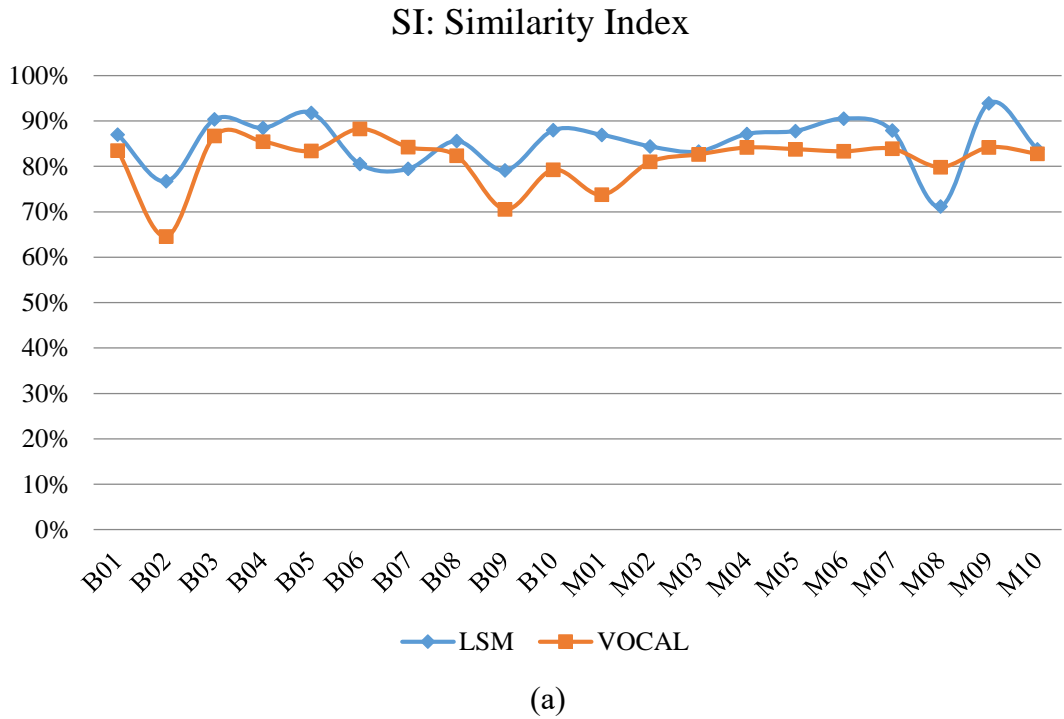

(a)

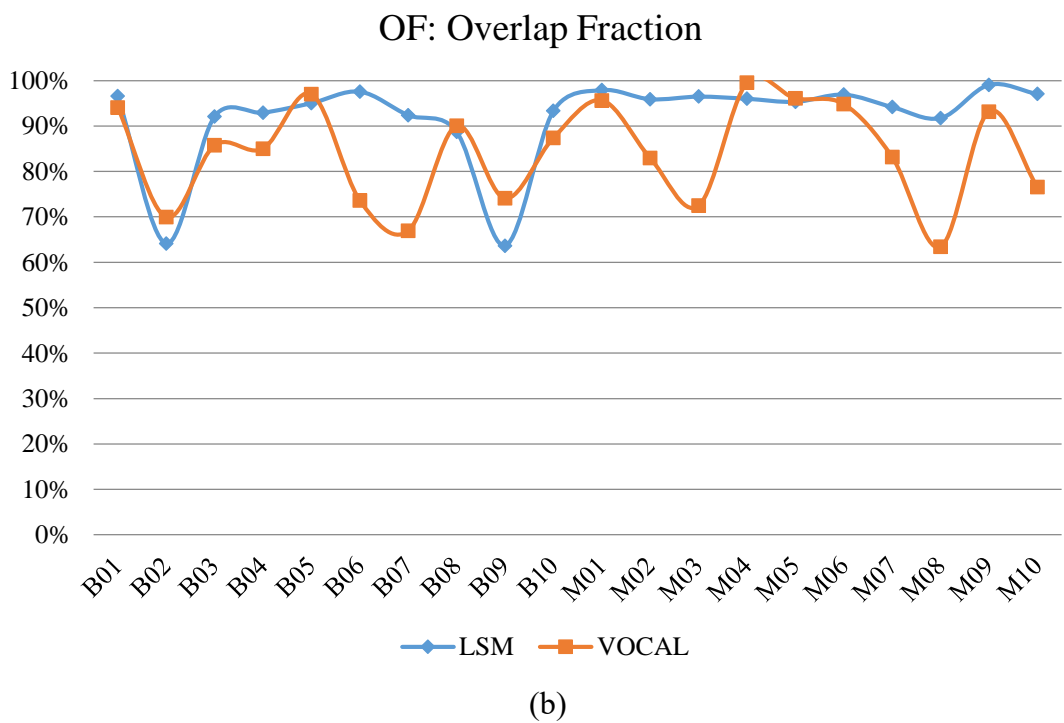

(b)

Figure 8. Similarity evaluation results of the assessment of the proposed LSM method and VOCAL method: (a) Similarity index (SI), (b) Overlap value (OV), (c) Overlap fraction (OF) and (d) Extra fraction (EF) (continued)

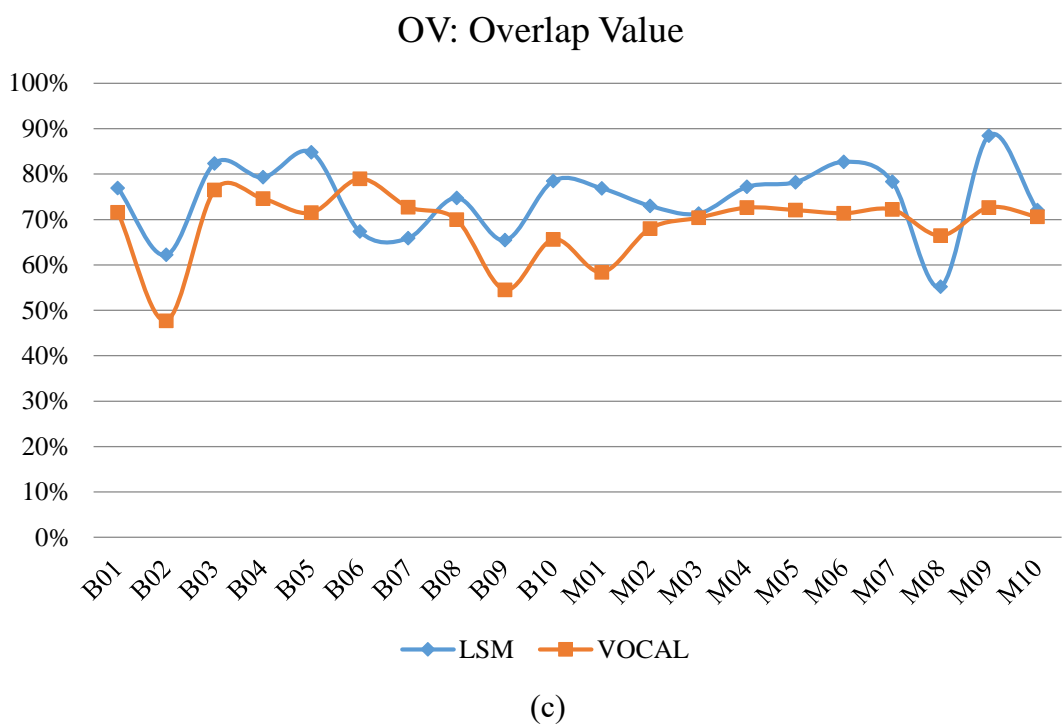

(c)

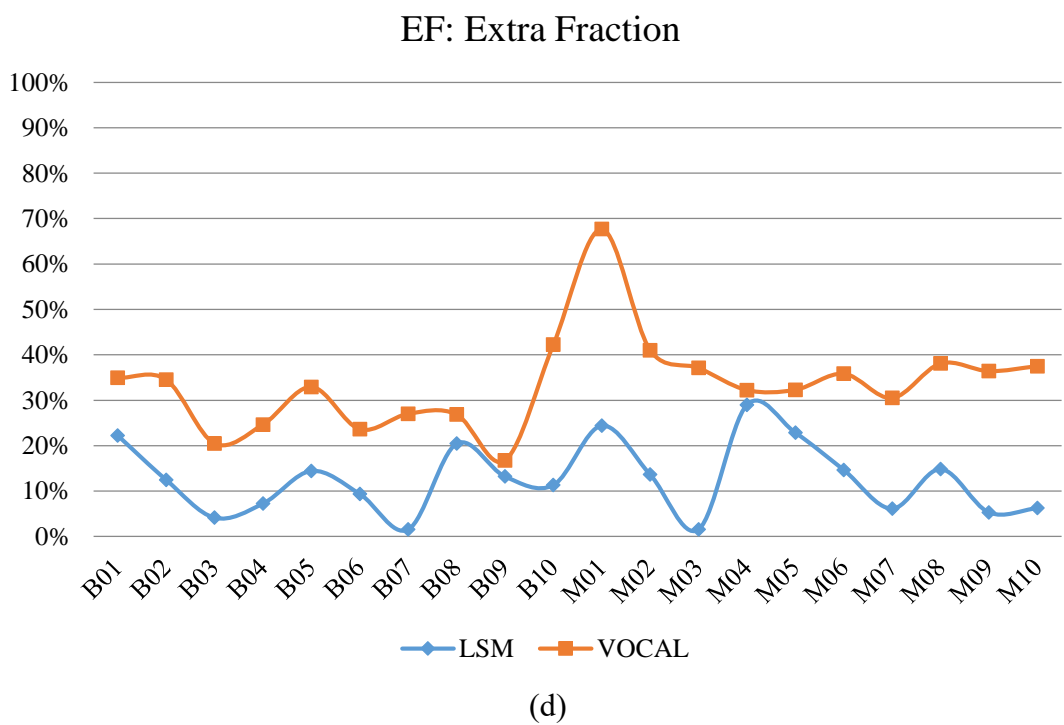

(d)

Figure 8. Similarity evaluation results of the assessment of the proposed LSM method and VOCAL method: (a) Similarity index (SI), (b) Overlap value (OV), (c) Overlap fraction (OF) and (d) Extra fraction (EF)